\documentclass[10pt,twocolumn,letterpaper]{article}

\usepackage[pagenumbers]{cvpr} 
\usepackage[dvipsnames]{xcolor}

\definecolor{cvprblue}{rgb}{0.21,0.49,0.74}
\usepackage[pagebackref,breaklinks,colorlinks,citecolor=cvprblue]{hyperref}
\usepackage{enumitem}

\usepackage{amsmath}
\usepackage{amsfonts}
\usepackage{amssymb}
\usepackage{epsfig}
\usepackage{graphicx}

\usepackage{booktabs}
\usepackage{multirow}

\newcommand{\T}{ICCC}

\def\paperID{3996} 
\def\confName{CVPR}
\def\confYear{2024}

\title{Learning by Correction: Efficient Tuning Task for Zero-Shot Generative Vision-Language Reasoning}

\author{
Rongjie Li\textsuperscript{\rm 1}\thanks{Equal Contribution.
This work was supported by NSFC 62350610269, 
the Shanghai Frontiers Science Center of Human-centered Artificial Intelligence, and the MoE Key Lab of Intelligent Perception and Human-Machine Collaboration (ShanghaiTech University). 
\quad ‡ denotes corresponding author. Code is available: \url{https://github.com/SHTUPLUS/ICCC_CVPR2024}
}
\quad Yu Wu\textsuperscript{\rm 1}\footnotemark[1]
\quad Xuming He\textsuperscript{\rm 1,2}\footnotemark[3] \\
\textsuperscript{\rm 1}School of Information Science and Technology, ShanghaiTech University \quad \\
\textsuperscript{\rm 2}Shanghai Engineering Research Center of Intelligent Vision and Imaging\\
\{lirj2, wuyu1, hexm\}@shanghaitech.edu.cn
}

\begin{document}
\maketitle

\begin{abstract}

Generative vision-language models (VLMs) have shown impressive performance in zero-shot vision-language tasks like image captioning and visual question answering. However, improving their zero-shot reasoning typically requires second-stage instruction tuning, which relies heavily on human-labeled or large language model-generated annotation, incurring high labeling costs. To tackle this challenge, we introduce Image-Conditioned Caption Correction (ICCC), a novel pre-training task designed to enhance VLMs' zero-shot performance without the need for labeled task-aware data. The ICCC task compels VLMs to rectify mismatches between visual and language concepts, thereby enhancing instruction following and text generation conditioned on visual inputs. Leveraging language structure and a lightweight dependency parser, we construct data samples of ICCC task from image-text datasets with low labeling and computation costs. Experimental results on BLIP-2 and InstructBLIP demonstrate significant improvements in zero-shot image-text generation-based VL tasks through ICCC instruction tuning.

\end{abstract}

\section{Introduction}\label{sec:intro}

\begin{figure}[t]
    \centering
    \includegraphics[width=1.0\linewidth]{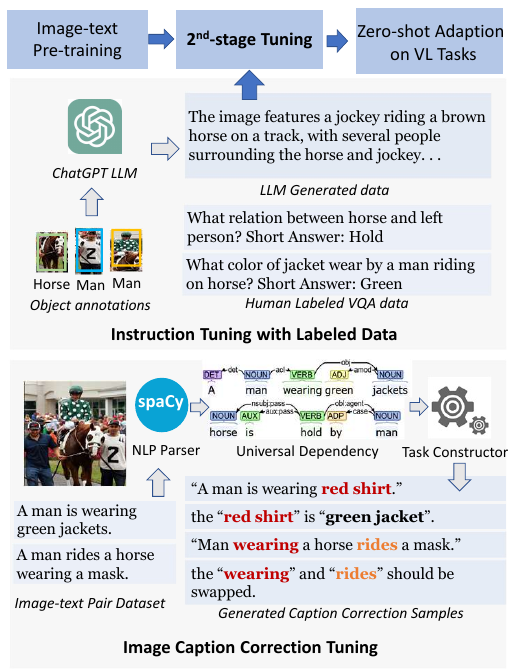}
    \vspace{-0.2cm}
    \caption{\textbf {Illustration of second-stage tuning for zero-shot VL task adaptation comparison.} The instruction tuning in recent works needs human label or LLM-generated data; in contrast, our image caption correction tuning is conducted on unlabeled image-text data with an NLP parser.}
    \label{fig:ad_fig}
    \vspace{-0.4cm}
\end{figure}

Vision-language models (VLMs) have demonstrated remarkable performance across a wide range of vision-language (VL) tasks, including image captioning~\cite{liu2023visual,li2023blip2,alayrac2022flamingo,wang2022ofa,yang2022unitab}, visual recognition~\cite{radford2021learning, zhou2022conditional, jia2021scaling}, image-text retrieval~\cite{radford2021learning,li2022blip}, and answering visual questions~\cite{liu2023visual,li2023blip2,wang2022ofa,yang2022unitab, alayrac2022flamingo}.
Generally, existing VLMs are able to conduct two essential tasks: image-text matching (ITM) and image-text generation (ITG).
The contrastive-based ITM aims to model the similarity between vision and text through a shared embedding~\cite{radford2021learning,jia2021scaling,li2022blip}.
In contrast, the generative-based ITG has more flexibility in adapting to various VL tasks.
Moreover, several recent VLM frameworks integrate the LLMs for ITG, which extend the powerful text generation across vision and text modality~\cite{li2023blip2,dai2023instructblip,alayrac2022flamingo,liu2023visual}, and allow VLMs to perform zero-shot inference on various VL tasks with impressive performance.

To perform zero-shot inference on VL tasks, the VLMs need to have generalizable text generation capability according to text inputs and concepts from the visual modality.
The existing works typically conduct second-stage instruction tuning for pre-trained VLMs with task-oriented data. This improves VLMs for following instructions to generate texts conditioned on visual modality, which ultimately enhances the zero-shot performance on VL-tasks such as InstructBLIP~\cite{dai2023instructblip} and LLAVA~\cite{liu2023visual}, as shown in the upper part of Fig.~\ref{fig:ad_fig}.
However, these methods necessitate substantial downstream task data annotation for fine-tuning, which is either human-labeled or generated externally by large language models. This process escalates labor costs throughout the system.

In this study, we introduce a novel pre-training task, Image-Conditioned Caption Correction (\T), aimed at enhancing VLMs' performance on zero-shot VL tasks. Our approach leverages the semantic dependency structure of language utilized for second-stage tuning of VLMs, using image-text data without task-specific annotation, as depicted in Fig.~\ref{fig:ad_fig}. 
By enforcing VLMs to identify and rectify mismatched concepts between visual and language, our method enhances VLMs' capability of generating text from the visual modality.
Importantly, the adopted universal semantic dependency~\cite{nivre2016universal} ensures comprehensive coverage of various concepts, including objects, their attributes, and interactions between them.
Furthermore, we construct the data from unlabeled image-text datasets only with a lightweight dependency parser, which achieves low labeling and computation costs.

Specifically, our pre-training framework first generates the task of image-conditioned text correction in an automatic manner. To this end, 
we develop a data construction pipeline with two components: the \textit{concept extractor} and the \textit{correction task constructor}.
Firstly, the \textit{concept extractor} identifies various concepts from the text modality.
With the off-the-shelf dependency parser, it extracts the set of language units by parsing the semantic dependency structure of text.
Subsequently, the \textit{correction task constructor} generates samples from the unlabeled image-text data according to language structure and concept set.
It swaps or replaces language units according to the extracted concept set, thereby creating concept-mismatched image-text pairs. 
Thanks to the universality of dependency, this approach allows us to create a wide variety of samples covering diverse visual-language concepts. 
The resulting text correction task requires VLMs to detect and recover the language units of mismatched concepts (words and phrases) according to the image.
Finally, we use the generated samples together with the original image-text data to fine-tune pre-trained VLMs with language modeling objectives.

We conduct extensive experiments on two VLMs, BLIP-2~\cite{li2023blip2} and InstructBLIP~\cite{dai2023instructblip}, and evaluate the zero-shot performance of ITG on the representative tasks: visual question answering and image caption.
Our findings reveal that our proposed method yields substantial improvements in zero-shot generalization based on the initial pre-train model without requiring any manually labeled or LLM-generated data.
The main contributions of our work are threefold:
\begin{itemize}[itemsep=0mm,topsep=0pt]
    \item We introduce a novel image-conditioned text correction fine-tuning strategy for VLMs that enhances their generalization of ITG for VL tasks.
    \item We developed an automated data construction pipeline that produces large amounts of samples for fine-tuning, all generated from image-text pairs without the need for human annotations or additional LLMs.
    \item We demonstrated notable improvements in the zero-shot generalization capabilities of VLMs across various VL tasks.
\end{itemize}

\section{Related Work}

\paragraph{Generative Vision Language Models}

With advancements in large-scale pretraining, vision-language models (VLMs) have demonstrated notable zero-shot generalization across various tasks. Unlike contrastive VLMs such as CLIP~\cite{radford2021learning} and ALIGN~\cite{jia2021scaling}, which focus on image-text similarity scores, generative VLMs output text based on image and text inputs for tasks like visual question answering and image captioning.

In earlier studies, generative VLMs utilized fusion-encoder transformers~\cite{tan2019lxmert,chen2020uniter,li2020oscar,zhang2021vinvl} to simultaneously encode visual and linguistic tokens, and subsequent models~\cite{cho2021unifying, wang2022ofa, yang2022unitab} integrated visual input information into different architecture language models, giving rise to unified generative VL transformer models.
Recently, with advancements in Large Language Models (LLMs)~\cite{zhang2022opt, chung2022scaling, zheng2023judging, ouyang2022training}, efforts have sought to utilize LLMs' capabilities to project visual input into the language embedding space. BLIP-2~\cite{li2023blip2} bridges the modality gap using a pre-trained Q-Former. Building upon BLIP-2, MiniGPT-4~\cite{zhu2023minigpt} and InstructBLIP~\cite{dai2023instructblip} focus on next-stage instruction tuning to further enhance performance. LLaVA~\cite{liu2023visual} involves GPT-4~\cite{openai2023gpt} in generating instructions and conversations for training. LLaMA-Adapter v2~\cite{gao2023llama} and LaVIN~\cite{luo2023cheap} employ adapters for architecture fine-tuning.

In our study, akin to LLaVA and Instruct-BLIP, we focus on leveraging the benefits of second-stage fine-tuning of pre-trained VLMs. However, unlike these approaches, we aim to achieve this without down-stream tasks data from human annotation and large models generation. Instead, we introduce a novel correction task, construct from the image-text data with light-weight pipeline.

\paragraph{Language Structure-based Data Augmentation}

Data augmentation is widely employed in machine learning, with common techniques such as mixup~\cite{zhang2017mixup}, CutMix~\cite{yun2019cutmix}, and RandAugment~\cite{cubuk2020randaugment} in computer vision, and back-translation~\cite{xie2020unsupervised}, random word editing~\cite{wei2019eda} in natural language processing.
In the field of vision and language (VL), an effective approach often involves synthesizing new data using pre-trained models as augmentation~\cite{cascante2023going}. For instance, BLIP~\cite{li2022blip} leverages the model's initially trained capabilities to generate additional training data, thereby further enhancing the model's performance.

Some researchers in visual-linguistic (VL) studies pursue performance enhancements on challenging tasks by crafting difficult negative samples. 
For instance, NegCLIP~\cite{yuksekgonul2022and} employs a language parser to swap elements between sentences, generating hard negative image-text pairs for contrastive fine-tuning.
SVLC~\cite{doveh2023teaching} achieves a similar outcome by substituting elements in sentences. 
Our approach leverages linguistic structure to improve image-text generation (ITG) tasks. Unlike image-text matching (ITM) tasks, which focus on learning similarity metrics, generative models encounter difficulties with negative samples. 
To address this, we introduce a correction task for instruction fine-tuning, using structural language information to construct data samples. This enhances the generalization performance of VLMs in zero-shot ITG-based visual language tasks.

\section{Preliminary}\label{sec:pre}

\paragraph{Image-text Generation of VLMs}
Image-text generation, a fundamental task of VLMs, involves generating text from both visual and textual inputs.
Specifically, VLMs with parameters $\mathbf{\Theta}_{vlm}$ generate output text from images and textual input in an auto-regressive manner.
The visual input is projected into language embedding as visual tokens $\mathbf{Z}$ together with textual input $\mathbf{w} = [w_1, w_2,..., w_{i-1}]$ and fed into the subsequent LLM to predict the next token $w_i$ of the sequence $\mathbf{w}$.
\begin{align}
w_i = \mathcal{F}_{vlm}( \mathbf{Z}, w_1, w_2,..., w_{i-1}; \mathbf{\Theta}_{vlm}).
\end{align}

\paragraph{Second-stage Tuning for Zero-shot Generation}

While generative image-text pre-training provides VLMs with aligned vision-language representations, effective instruction-following of image-text generation for diverse zero-shot VL tasks is crucial.
Recent works address this issue by leveraging either human-labeled~\cite{dai2023instructblip} or large language model-generated task-oriented data~\cite{liu2023visual} for instruction tuning.
Concretely, in the second stage of tuning, the VLMs are optimized with the same objective of generative image-text pre-training with task-oriented data:
\vspace{-0.4cm}
\begin{align}
\mathop{\arg\max}_{\mathbf{\Theta}_{vlm}} \sum_{i=1}^{K} \log P(w_i| \mathbf{Z}, w_1, w_2,..., w_{i-1}; \mathbf{\Theta}_{vlm}).
\vspace{-0.7cm}
\end{align}

\begin{figure*}[t]
    \centering
    \includegraphics[width=0.99\linewidth]{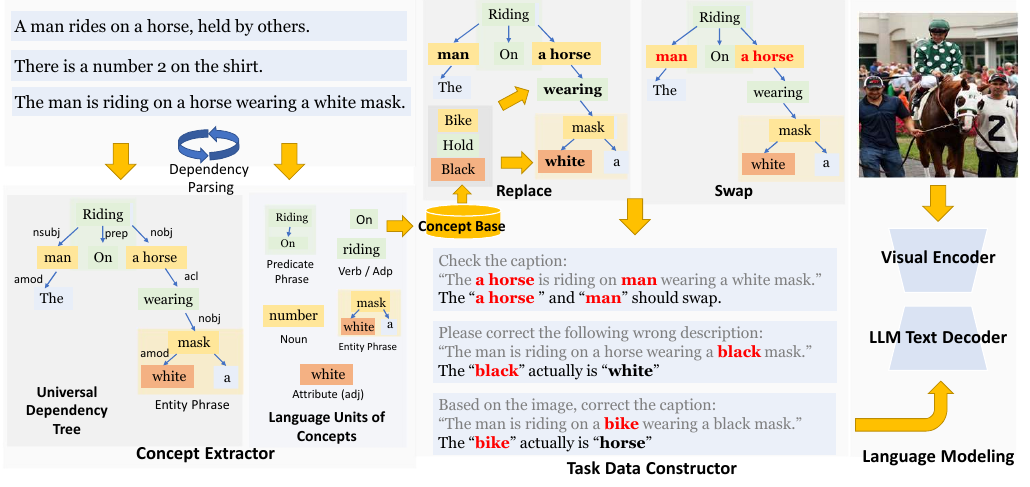}
    \vspace{-0.4cm}
    \caption{\textbf {Illustration of the overall pipeline of \T.} The concept extractor parses the sentence to obtain linguistic units of concepts. The task data constructor aims to produce the sample according to the sentence structure with the ``{\textsf{\small{replace}}}'' and ``\textsf{\small{swap}}'' operations. Finally, the generated \T~data is used for image-to-text generative training for VLMs.}
    \label{fig:main_fig}
    \vspace{-0.4cm}
\end{figure*}

\section{Our Approach}\label{sec:method}

In this section, we introduce our proposed Image Conditional Caption Correction (\T) task for second-stage tuning, which constructs data from unlabeled data with low resource consumption.
We first provide a \textit{Task Definition} of \T~in Sec.~\ref{sec:task}.
Then we introduce the data construction pipeline for \T, which includes two modules: the \textit{Concept Extractor} and the \textit{Correction Data constructor}.
The \textit{concept extractor} parses the text structure and extracts the concept set for the following data construction, discussed in Sec.~\ref{sec:concept}.
Subsequently, the \textit{correction data constructor} generates data by augmenting text structure with the extracted concept set, described in Sec.\ref{sec:constructor},
Finally, Sec.~\ref{sec:learning} outlines our training and inference procedures for incorporating our task into VLM pre-training.

\subsection{Task definition}\label{sec:task}

As illustrated in Fig.~\ref{fig:main_fig}, the \T~task involves identifying and correcting language units of mismatched concepts of caption.
To define the \T~task, we need to define the concept set, how we perturb to produce concept-mismatched samples, and how the perturb operation changes the language structure.

For each sentence, concepts are composed of linguistic units representing their semantic meaning.
These linguistic units are categorized into five types based on semantics and language granularity: $\mathcal{E}=\{$\textit{entity phrase, predicate phrase, attribute phrase, noun word, verb word}$\}$.
The \textit{entity phrase}, \textit{noun word}, \textit{attribute phrase} represent the object-level semantic, and \textit{predicate phrase}, \textit{verb word} represent the composition-level semantic.
The \textit{phrase} and \textit{word} represent different levels of language granularity.

There are two operations for perturbing the language structure: $\{\textsf{\small{replace}}, \textsf{\small{swap}}\}$.
The \textsf{\small{replace}} involves substituting a concept with another one of the same type from the image-text pair dataset, while the \textsf{\small{swap}} involves swapping the positions of two concepts of the same type within the original caption. More details are introduced in Sec.~\ref{sec:constructor}.
Consequently, replace focuses on modeling the intrinsic meaning of individual concepts, while swap prioritizes the order of compositional concepts within sentences.
Overall, the task is to use linguistic unit modifications of image-text pairs, providing universal concepts aligned between vision and language, which improves text generation capabilities across two modalities.

\subsection{Concept Extractor}\label{sec:concept}

As mentioned earlier, the data of \T~is generated by perturbing concepts in $\mathcal{E}$ according to the structure of sentences.
The concept extractor takes captions as input and parses the dependency to identify linguistic units of concepts, which the following task constructor then uses to generate data samples.
The concept extractor module has three processes:
(1) universal dependency parsing; (2) linguistic unit selection and grouping; (2) concept collection.

\paragraph{Universal Dependency Parsing}
We use the off-the-shelf universal dependency parser implemented by the spaCy~\cite{spacy2} software library to extract the dependency structure of the original caption.
As shown in Fig.~\ref{fig:main_fig}, the dependency parser translates the sentence into a universal dependency tree~\cite{nivre2016universal}.
The node of the dependency tree is the minimal linguistic unit $u_{[pos]}$, which has Part-of-Speech (POS) tags to indicate its grammatical role.
The edge represents the dependency relation between linguistic units $r_{[REL]}$, which also has the relation type (REL).
The dependency structure provides us with an indication of how linguistic units organize to represent the concept of a sentence.

\paragraph{Linguistic Unit Selection and Grouping}
We extract the linguistic unit set of concept $U_{\textit{concept}}$ within $\mathcal{E}$ by grouping and selecting $u_{[pos]}$.
To achieve this, we design a heuristic method by traversing the dependency tree to select the corresponding $u_{[pos]}$.
Firstly, we extract the concept represented by the concept of a word-level linguistic unit according to the POS tag:
\begin{enumerate}
  \item $U_{\textit{noun word}}$ :  $u_{[noun]}$
  \item $U_{\textit{verb word}}$ :  $u_{[verb]}$ 
\end{enumerate}
Based on such word-level concepts $u_{[pos]}$, we further extract linguistic units of phrase-level concepts, such as \textit{entity phrase, predicate phrase, attribute phrase}, by grouping the $u_{[pos]}$ into $U_{\textit{concept}}$ according to the $r_{[REL]}$ associated with word-level concept units:
\begin{enumerate}
  \item $U_{\textit{entity phrase}}$: $u_{[adj]}$ and $U_{[det]}$ adjacent to $U_{[noun]}$
  \item $U_{\textit{predicate phrase}}$: all $u_{[pos]}$ within two $u_{[noun]}$
  \item $U_{\textit{attribute phrase}}$: all $u_{[amod]}$ adjacent to $u_{[noun]}$
\end{enumerate}
To this end, we extract the linguistic units of each concept type $U_{\textit{concept}}$ from the text according to the dependency structure.

\paragraph{Concept Collection}
For each sentence from the image-text dataset, we collect all $U_{\textit{concept}}$ according to their concept type, respectively.
We merge the $U_{\textit{noun word}}$ with the $U_{\textit{entity phrase}}$ and the $U_{\textit{verb word}}$ with the $U_{\textit{predicate phrase}}$ into the same concept type, respectively.
This global-level concept base stores all the concepts that occurred in the dataset. We filter them by frequency to remove infrequent concepts, which could be extracted from low-quality captions or parsing errors, and the most frequent, which could be a trivial concept or language bias that occurred uniformly.

\subsection{Correction Task Data Constructor}\label{sec:constructor}
The correction task data constructor uses the extracted concept structure to generate data samples with multi-level concept mismatch captions.
The data constructor takes the input text, which has identified concepts of language units and concept-based units, and produces the augmented text with expected corrections.

Specifically, the data constructor perturbs the initial linguistic structure by predefined operations: \textsf{\small{replace}}, \textsf{\small{swap}}, as shown in Fig.~\ref{fig:main_fig}.
Specifically, the \textsf{\small{replace}} randomly selects the concept of the original text and replaces it with an external $U_{\textit{concept}}$ that has the same semantic type but does not occur in the current text from the concept base.
For instance, the \textit{entity phrase} can be replaced by all object-level concepts, such as \textit{noun word} or \textit{entity phrase}.
The \textsf{\small{swap}} operation, instead of replacing concept by concept, swaps the two linguistic unit sets with the same concept type in the text.
This perturbation operation provides the order mismatch of the concept.
To calibrate those two operations, we randomly select one by Bernoulli distribution with a preset parameter $p_s \in [0,1]$ to decide which type of operation is used for perturbation. In those cases where a sentence doesn't have two of the same concept-type linguistic units, we simply use the replace operation.

After the perturb operation, we map back the tree structure sentence into sequence and construct the caption correction samples, which are composed of the following components:
correction instruction prompt, perturbed caption, and correcting description, as shown in Fig.~\ref{fig:main_fig}.
The correction instruction prompt is selected from a template base, and the correcting description includes the mismatch concept and the correct ones.

\subsection{Training and Inference}\label{sec:learning}
The data generated by ICCC is utilized in the second-stage tuning process to improve VLMs, as introduced in Sec~\ref{sec:pre}. 
We combine ICCC data samples with original image-text pairs for training to prevent focusing too much on the specific task and catastrophic forgetting. 
The proportion of ICCC data samples to original pairs in each training batch is determined by a hyper-parameter $p_c$.
The overall pre-training objective remains unchanged from the initial image-text generation pre-training. 
This design enables VLMs to grasp the alignment of concepts and facilitate instruction following for various downstream generative VL tasks, enhancing their performance across different tasks.
During inference, VLMs conduct standard image-to-text generation for VL tasks, consistent with previous generative VLMs~\cite{li2023blip2, dai2023instructblip}.
\begin{table*}[ht!] 
\centering
\resizebox{0.67\linewidth}{!}{
{\fontsize{10}{12}\selectfont
\begin{tabular}{l|cccc|ccc}\toprule
\multirow{2}{*}{\textbf{BLIP-2}} & \multirow{2}{*}{\textbf{GQA}} & \multirow{2}{*}{\textbf{OK-VQA}} & \multirow{2}{*}{\textbf{VQAv2}} & \multirow{2}{*}{\textbf{VSR}} & \multicolumn{3}{c}{\textbf{NoCaps} }    \\  
      &   &  &    &  & B@4     & S     & C        \\ \midrule
OPT2.7B          &  33.5                  & 26.6                   & 51.9      &    \textbf{48.3}     & 43.6  & 13.8  & 105.7 \\ 
OPT2.7B w/ \T        &   \textbf{38.9}                 & \textbf{29.5}     & \textbf{54.3}    &   47.8      &    \textbf{46.0}	& \textbf{14.3}	& \textbf{111.9} 
\\ \midrule
OPT6.7B          &  35.5                & 30.7                  & 52.6  &   48.5     & 41.5 & 13.0 & 101.4 
\\
OPT6.7B w/ \T       &  \textbf{38.2}                & \textbf{31.7}    & \textbf{58.8}   &    \textbf{51.5}     & \textbf{44.1} & \textbf{13.6} & \textbf{106.9} 
\\ \midrule
FlanT5-XL         &  44.0                  & 40.7     &  63.1   &    63.4   & 42.2  & 13.3 & 103.1 
\\
FlanT5-XL w/ \T      & \textbf{44.6}	              & \textbf{41.0}        & \textbf{64.0}  &   \textbf{64.2}       & \textbf{43.9} & \textbf{13.6} & \textbf{106.0} 
\\ 
\bottomrule
\end{tabular}}}
\caption{\textbf{The zero-shot evaluation results on BLIP-2 experiments.} For three VQA datasets, we report top-1 accuracy (\%) on the testdev set of GQA~\cite{hudson2019gqa}, VSR~\cite{liu2023vsr}, the test set of OK-VQA~\cite{marino2019ok}, and the validation set of VQAv2~\cite{goyal2017making}. For IC, we report metrics of BLUE@4 (B@4), CIDEr (C), and SPICE (S) on the validation set of NoCaps.}\label{tab:blip2}
\vspace{-0.1cm}
\end{table*}

\begin{table}[h] 
\centering
\resizebox{0.8\columnwidth}{!}{
\begin{tabular}{l|ccc}\toprule
{\textbf{Model}} & {\textbf{GQA}} & {\textbf{VSR}} & {\textbf{NoCaps}}           \\  \midrule
InstructBLIP                 & 48.4                 & 61.1            &   14.2 \\
InstructBLIP w/ \T          & \textbf{49.8}                 & \textbf{63.1}             &   \textbf{15.7}  \\    
\bottomrule
\end{tabular}}
\caption{\textbf{The zero-shot evaluation results on InstructBLIP experiments.} We report the SPICE score for NoCaps.}\label{tab:instructblip}
\vspace{-0.2cm}
\end{table}

\section{Experiments}

In this section, we thoroughly explore the effectiveness of our proposed training task. We detail the experimented models and implementation in Sec.~\ref{sec:exp_config}, conduct primary zero-shot evaluations in Sec.~\ref{sec:exp_zero}, present ablation studies in Sec.~\ref{sec:exp_abl}, and offer qualitative insights in Sec.~\ref{sec:exp_qua}.

\subsection{Experiments Configuration} \label{sec:exp_config}
\paragraph{Models and Training Data}

In the realm of generative VLMs, we select BLIP-2~\cite{li2023blip2} and InstructBLIP~\cite{dai2023instructblip}, representing different pre-training data paradigms.
BLIP-2 utilizes image-text pair data without instructions, employing a frozen image encoder and a large language model. It integrates a lightweight querying transformer for mapping visual information into the language embedding space.
Instead, InstructBLIP incorporates instruction tuning, enhancing the architecture of pre-trained BLIP-2 with an instruction-aware visual feature extractor. It undergoes complementary training with additional task-specific instruction-tuning data.

In our model setup, we explore three variants of BLIP-2 to assess the generality of our training method across different LLM architectures. They share the same image encoder (ViT-G/14 from EVA-CLIP~\cite{fang2023eva}) but employ distinct frozen LLMs: OPT~\cite{zhang2022opt} with 2.7B and 6.7B parameters, and FlanT5-XL~\cite{jia2021scaling} with 3B parameters.
For InstructBLIP, we experiment with the Vicuna-7B~\cite{zheng2023judging} version. Initializing both models with pre-trained parameters, we freeze the vision encoder and LLM, focusing solely on training their Q-former and fully connected projection network.

We construct \T~samples using our proposed method on the COCO Caption~\cite{chen2015microsoft} and Visual Genome (VG) Caption~\cite{krishna2017visual} datasets, totaling approximately 1 million image-text pairs. These samples are then utilized for language modeling training and image captioning in a second-stage training paradigm outlined in Sec.~\ref{sec:learning}.

\paragraph{Implementation Details}
We implement and evaluate our method using the LAVIS library~\cite{li-etal-2023-lavis}, and mainly followed the training setup for the original models. The AdamW~\cite{loshchilov2018decoupled} optimizer with $\beta_1 = 0.9$, $\beta_2 = 0.999$, and a weight decay of $0.05$ was employed. We used a linear warmup of the learning rate over the initial 1,000 steps, increasing from $10^{-8}$ to $10^{-5}$, followed by a cosine decay with a minimum learning rate of $0$. Batch sizes varied across models: 64 for BLIP-2 OPT2.7B and BLIP-2 FlanT5-XL, 28 for BLIP-2 OPT6.7B, and 24 for InstructBLIP. The images are resized to size 224$\times$224, and we apply random resized cropping and horizontal flipping augmentations. All training spanned a maximum of 20,000 iterations, with model performance validated every 1,000 iterations. Each training process utilized four Nvidia A40 (40G) GPUs, completed within a day.

Regarding the hyperparameters, we conducted comparison experiments. We set $(p_c, p_s)$ as $(0.3, 0.15)$, $(0.3, 0)$, $(0.01, 0.2)$, and $(0.3, 0.3). $ for BLIP-2 OPT2.7B, BLIP-2 OPT6.7B, BLIP-2 FlanT5-XL, and InstructBLIP, respectively. Further details will be demonstrated in Sec.~\ref{sec:exp_abl}.

\subsection{Zero-shot Evaluations} \label{sec:exp_zero}

We present zero-shot evaluation results in Tab.~\ref{tab:blip2} and Tab.~\ref{tab:instructblip} for BLIP-2 and InstructBLIP experiments. Our second-stage training strategy \T~consistently improves zero-shot performance across various tasks and datasets, like visual question answering (VQA) and image captioning (IC), demonstrating enhanced zero-shot generalization.

The experimental results on BLIP-2 (Tab.~\ref{tab:blip2}) demonstrate the effectiveness of \T~on models pre-trained by image-text pairs. Notable findings include:

\begin{itemize}
\item
\T~yields consistent improvements for BLIP-2 OPT2.7B, BLIP-2 OPT6.7B, and BLIP-2 FlanT5-XL.
Specifically, the BLIP-2 OPT2.7B with \T~shows a significant enhancement in GQA (+5.4\%), while the BLIP-2 OPT6.7B improves considerably in VQAv2 (+6.2\%). These results emphasize \T's ability to refine vision-conditioned language generation, irrespective of LLM sizes and architectures.

\item
In captioning tasks, SPICE metrics highlight improved accuracy at the structured scene level, including VL relations and attributes. Our task design, focusing on learning ITG generalizations for various concept roles, not only enhances object recognition accuracy but also improves understanding of object attributes and relations.

\item
In the InstructBLIP experiment results shown in Tab.~\ref{tab:instructblip}, we note a consistent improvement in zero-shot evaluation benchmarks, even when the original model heavily relies on instruction tuning. This indicates the effectiveness of our approach across diverse pre-training image-text data, offering a valuable complementary method for augmenting visual-linguistic knowledge.

\item
We emphasize the Visual Spatial Reasoning (VSR) benchmark~\cite{liu2023vsr}, created to evaluate a model's grasp of spatial relationships for vision-language reasoning. These enhancements stem from the varied concept data samples in our \T~pre-training, which are able to encompass a broad spectrum of vision-language relational concepts.

\item
Additionally, we assess the effectiveness of \T~on another VLM (LLAVA~\cite{liu2023visual}) and various vision-language reasoning datasets, including ScienceQA-IMG~\cite{saikh2022scienceqa}, MM-VET~\cite{yu2023mm}, and hallucinations~\cite{rohrbach2018object}. Further experimental results can be found in the supplementary material.
\end{itemize}

\begin{figure*}[tbp]
  \centering
  \includegraphics[width=1.0\linewidth]{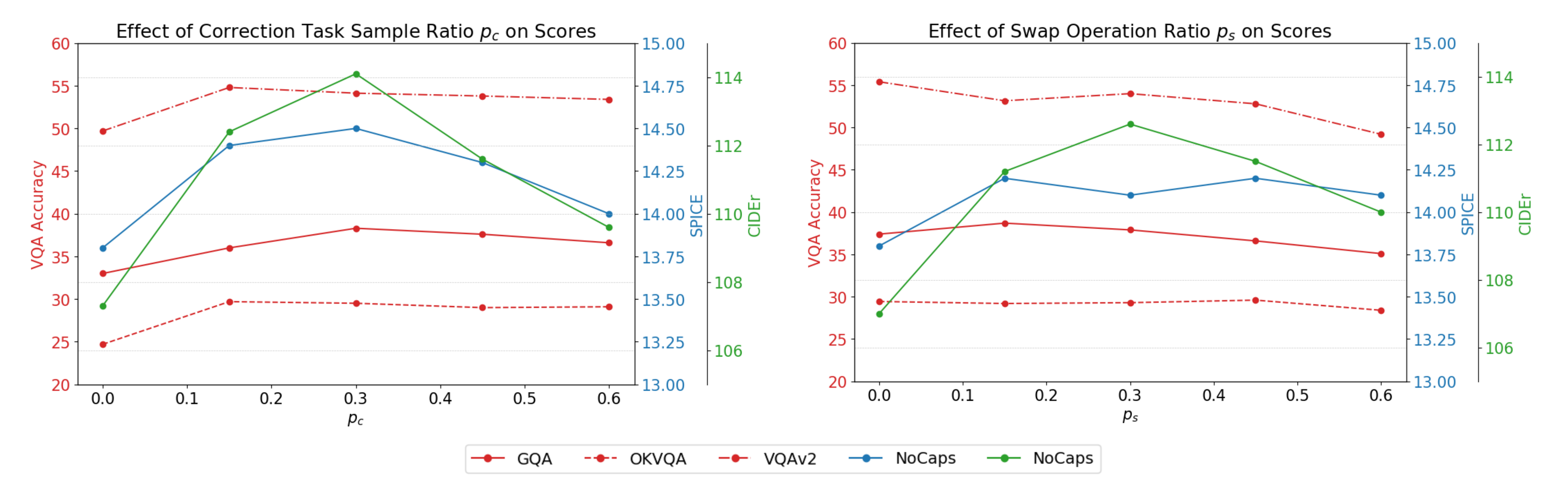} 
  \vspace{-0.5cm}
  \caption{\textbf{Hyper-parameter searching on $p_c$ and $p_s$.}}
  \label{fig:p_c}
  \vspace{-0.5cm}
\end{figure*}

\subsection{Ablation Study} \label{sec:exp_abl}

\begin{table}[t]
\resizebox{\columnwidth}{!}{
\begin{tabular}{l|ccc|ccc|ccc}
\toprule
& \multicolumn{3}{l|}{\textbf{Data Source}}                                                                               & \multirow{2}{*}{\textbf{GQA}} & \multirow{2}{*}{\textbf{OK-VQA}} & \multirow{2}{*}{\textbf{VQAv2}} & \multicolumn{3}{c}{\textbf{NoCaps}} \\ 
 & C+V                & VQAv2                     & w/ \T      &                      &                        &                       & B@4     & S       & C      \\ \midrule
1 & -                                                                & -                         & -                         & 33.5                 & 26.6                   & 51.9                  & 43.6   & 13.8   & 105.7  \\ \midrule
2 & \checkmark  & -                         & -                         & 33.0                 & 26.3                  & 49.7                     & 45.6   & 13.9   & 108.1  \\
3 & \checkmark                  & -                         & \checkmark &  \textbf{38.9}                 & \textbf{29.5}                 & \textbf{54.3}             &    \textbf{46.0}	& \textbf{14.3}	& \textbf{111.9}   \\  \midrule
4 & \checkmark                  & \checkmark & -                         & 41.3                 & 38.0                   & -                       & \textbf{45.2}   & 12.6   & 106.4  \\
5 & \checkmark & \checkmark & \checkmark & \textbf{42.8}                 & \textbf{39.1}                   & -                   & 45.0   & \textbf{14.0}   & \textbf{108.2} \\ \bottomrule
\end{tabular}}
\vspace{-0.2cm}
\caption{\textbf{Ablation study on our method upon different second-stage training data types.} C+V represents COCO and VG caption data. Since our focus is on zero-shot performance, we omit the results on the VQAv2 validation set after training with the VQAv2 training set.}\label{tab:data}
\vspace{-0.4cm}
\end{table}

\paragraph{Effectiveness of Correction Task} 

In Tab.~\ref{tab:data}, we compare our approach with alternative second-stage training methods, focusing on the BLIP-2 OPT2.7B model. We explore two types of second-stage training: one utilizing only COCO and VG caption datasets (lines 2 and 3), and the other incorporating the VQAv2 training set (lines 4 and 5) to assess the impact of instruction tuning. Our approach exclusively targets caption datasets, maintaining consistent task sample construction parameters.

Experimental results indicate that our training approach consistently outperforms conventional second-stage methods, regardless of the inclusion of VQA data for instruction-tuning. This underscores the efficacy of our task, emphasizing its superiority over traditional ITG second-stage training strategies.


\paragraph{Ablation for Task Construction}

In Tab.~\ref{tab:concept}, we demonstrate the necessity of including different types of concepts to edit for constructing mismatched captions.
The experiments were conducted on the BLIP-2 OPT2.7B model, maintaining consistent hyper-parameters. Results indicate that using concepts of different natures for task construction yields complementary effects.

\begin{itemize}
\item
In the first part, we investigate influence of language granularity, focusing on tasks constructed around words and phrases.
As shown in lines 3–4 of Tab.\ref{tab:concept}, our findings reveal the significance of phrase-level concepts for the VQA task, while word-level concepts aid in refining the model's understanding of individual words, thereby improving captioning accuracy metrics such as BLUE@4 and CIDEr.

\item
In the subsequent section (lines 5–6), we delve into the roles of various concepts in semantics, exploring the impact of tasks centered on entity-level, relation-level, and attribute-level concepts.
Referencing Tab.\ref{tab:concept}, our analysis reveals that focusing solely on learning at the entity level yields strong performance in VQA tasks but may sacrifice accuracy in comprehending relations and attributes, thereby resulting in sub-optimal image captioning performance.

\item
We conducted an ablation study to assess the impact of instructions and language structure (line 7, Tab.\ref{tab:concept}). We retained the instructions of the ICCC task while randomly altering words within the sentences. Surprisingly, the addition of correction instructions and random language masking did not enhance the generalization capability of VLMs on downstream tasks. This highlights the importance of language structure in instruction tuning.

\end{itemize}

In conclusion, the diversity in concept extraction allows our method to perform well on various zero-shot generation tasks, demonstrating strong generality.

\paragraph{Hyper-parameter Selection}

\begin{table}[t]
\resizebox{\columnwidth}{!}{
\begin{tabular}{l|l|ccc|ccc}
\toprule
\multicolumn{1}{c|}{\multirow{2}{*}{}} & \multirow{2}{*}{\textbf{Type}} & \multirow{2}{*}{\textbf{GQA}} & \multirow{2}{*}{\textbf{OK-VQA}} & \multirow{2}{*}{\textbf{VQAv2}} & \multicolumn{3}{c}{\textbf{NoCaps}} \\
\multicolumn{1}{c|}{}                               &                            &                      &                        &                        & B@4      & S      & C      \\ \midrule
1    & none                        & 33.0                 & 26.3                 & 49.7             &    45.6	& 13.9	& 108.1      \\
2    & all                        & \textbf{38.9}                 & 29.5                 & 54.3             &    46.0	& 14.3	& 111.9      \\ \midrule
3                    & \textit{noun}, \textit{verb}   
& 36.4                & \textbf{30.7}                  & 52.1          &     46.0 & 14.3	& \textbf{114.3}       \\
  4    & \textit{ent}, \textit{pred}, \textit{attr}
  & 38.3                 & 27.5                   & \textbf{54.9}                   & 45.2 &	\textbf{14.7} &	111.3   \\ \midrule
5   &   \textit{noun}, \textit{ent}     
& 38.5                 & 29.8                   & 54.8             &   46.5 &	14.4 &	112.4     \\
   6  & \textit{verb}, \textit{pred}, \textit{attr}         
   & 36.0                 & 29.6                   & 54.1                   & \textbf{46.9}	& 14.5 &	114.1 \\ \midrule
   7  & random  & 33.0                & 26.2                   & 48.6                   & 44.1	& 14.5 &	107.1 \\ \bottomrule
\end{tabular}}
\vspace{-0.1cm}
\caption{\textbf{Ablation study on editing different subsets of concept types for mismatched caption construction.}}\label{tab:concept}
\vspace{-0.48cm}
\end{table}

The hyperparameters $p_c$ and $p_s$ are pivotal in adjusting the distribution of task samples in our approach. We conducted a comprehensive hyperparameter search for each experimental model, analyzing the impact of these parameters on model performance. Fig.~\ref{fig:p_c} illustrate the performance variations of BLIP-2 OPT2.7B after second-stage training under different hyperparameter settings:

\begin{itemize}
\item
In the left of Fig.~\ref{fig:p_c}, with $p_s$ fixed at 0.1, we study the influence of different correction task sample proportions by varying $p_c$ at intervals of 0.15.

\item
In the right of Fig.~\ref{fig:p_c}, with $p_c$ fixed at 0.3, we explore the impact of different \textsf{\small{swap}} operation proportions on mismatched caption sampling by varying $p_s$ at intervals of 0.15. 

\end{itemize}

In our experiments, we present several key findings:

\begin{itemize}
\item From both figures, it is evident that the impact of hyperparameters $p_c$ and $p_s$ on model performance follows a consistent trend, initially increasing before decreasing. This phenomenon arises due to the risk of language bias when incorporating excessive correction task samples, particularly those generated through \textsf{\small{swap}} operations.

\item Notably, the saturation point of $p_s$ is reached early, suggesting that the model can easily identify and correct mismatched captions created through \textsf{\small{swap}} operations. Despite this, proper inclusion of such samples proves beneficial for overall performance enhancement, particularly in image captioning tasks.

\item Furthermore, our investigation reveals that BLIP-2 FlanT5-XL tends to overfit on our task samples, possibly attributed to its encoder-decoder architecture. This results in the Q-former learning an overfitted encoding of soft prompts associated with task text patterns. Nonetheless, even minimal incorporation of correction task samples ($p_c=0.01$) effectively serves our training objectives while mitigating the risk of overfitting.
\end{itemize}

\subsection{Qualitative Results} \label{sec:exp_qua}
\begin{figure}[tbp]
  \centering
  \includegraphics[width=1.0\linewidth]{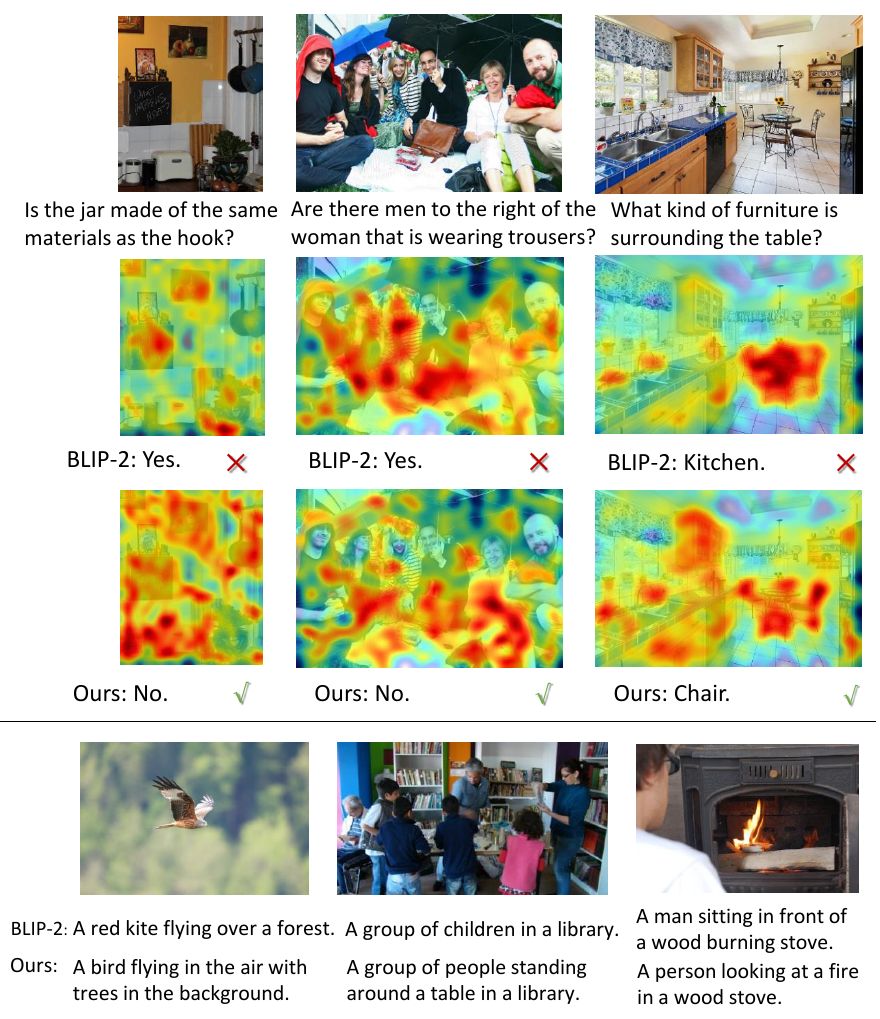} 
  \vspace{-0.4cm}
  \caption{
  \textbf{Visualization results include model output examples and attention gradients on images.} The first block illustrates three examples from the GQA testdev set, while the second block showcases three examples from the NoCaps validation set. With our training, the model demonstrates improved accuracy in focusing on prompt-relevant image regions. Additionally, it generates captions with more detailed descriptions of scenes and actions.
  }
  \label{fig:vis}
  \vspace{-0.4cm}
\end{figure}

To provide a more intuitive understanding of the effects of our training, we show some qualitative model output examples in Fig.~\ref{fig:vis}. We demonstrate the differences between the model outputs and intermediate results on the VQA and IC tasks before and after our second-stage training respectively. The heat maps show the gradient of the output logit corresponding to the ground truth token with regard to the image attention, which reflects the potential visual contribution of the model output.

According to the outputs for VQA, the results reveal that the proposed correction task training produces interpretable improvements to VQA responses. Our model demonstrates increased accuracy in locating objects referenced in prompts and exhibits a more precise understanding of compositional concepts, avoiding biases toward prominent objects. For instance, in the first example, our model focuses on the ``hook", which the original model misses. In the second example, our model correctly considers the composed concept of ``the woman that is wearing trousers". Lastly, our model interprets the relational concept ``surrounding the table" and directs attention appropriately toward the relevant region, rather than the table itself. Overall, our findings indicate that this training methodology enhances VLMs' capacity for interpreting corresponding visual concepts expressed through language.

Examples from our experiments with IC illustrate this improved ability to correct misidentified objects and provide richer, more fine-grained descriptions of object relationships and actions. For instance, our model correctly identifies the wrong ``red kite" concept in the first case and provides more nuanced details about the scene, such as ``standing around a table" and ``looking".
We posit that the primary source of improved overall captioning performance lies in the generation of more accurate and comprehensive concepts in captions.

Conclusively, our study indicates that the proposed training method potentially enhances generative VLMs' capacity for aligning concepts across multiple types and granularities in visual and language modalities. We believe this insight could shed light on the future development of cost-effective, multi-granularity, and structured generative VL pre-trained models.

\section{Conclusion}

In this work, we propose the image-conditioned text correction task for enhancing zero-shot text generation with unlabeled data.
In this task, VLMs requires to identify and correct the error in accordance with the vision modality via text generation.
We propose a scalable and cost-effective data construction framework for generating the image-text pair for this task by utilizing the inherent structure of language.
The experimental results indicate that the implementation of our training framework substantially improves the ability of VLMs to generalize across a range of VL tasks involving image-to-text generation.

\noindent \textit{Discussion of Limitation} Due to the limitation of the computing resource, extending our approach to larger datasets and models remains unexplored. Future work should explore its application to more extensive datasets and diverse large vision-language models.

{
\small
\bibliographystyle{ieeenat_fullname}
\bibliography{main}

\begin{thebibliography}{45}
\providecommand{\natexlab}[1]{#1}
\providecommand{\url}[1]{\texttt{#1}}
\expandafter\ifx\csname urlstyle\endcsname\relax
  \providecommand{\doi}[1]{doi: #1}\else
  \providecommand{\doi}{doi: \begingroup \urlstyle{rm}\Url}\fi

\bibitem[Alayrac et~al.(2022)Alayrac, Donahue, Luc, Miech, Barr, Hasson, Lenc,
  Mensch, Millican, Reynolds, et~al.]{alayrac2022flamingo}
Jean-Baptiste Alayrac, Jeff Donahue, Pauline Luc, Antoine Miech, Iain Barr,
  Yana Hasson, Karel Lenc, Arthur Mensch, Katherine Millican, Malcolm Reynolds,
  et~al.
\newblock Flamingo: a visual language model for few-shot learning.
\newblock \emph{Advances in Neural Information Processing Systems},
  35:\penalty0 23716--23736, 2022.

\bibitem[Cascante-Bonilla et~al.(2023)Cascante-Bonilla, Shehada, Smith, Doveh,
  Kim, Panda, Varol, Oliva, Ordonez, Feris, et~al.]{cascante2023going}
Paola Cascante-Bonilla, Khaled Shehada, James~Seale Smith, Sivan Doveh,
  Donghyun Kim, Rameswar Panda, Gul Varol, Aude Oliva, Vicente Ordonez, Rogerio
  Feris, et~al.
\newblock Going beyond nouns with vision \& language models using synthetic
  data.
\newblock In \emph{Proceedings of the IEEE/CVF International Conference on
  Computer Vision}, pages 20155--20165, 2023.

\bibitem[Chen et~al.(2015)Chen, Fang, Lin, Vedantam, Gupta, Doll{\'a}r, and
  Zitnick]{chen2015microsoft}
Xinlei Chen, Hao Fang, Tsung-Yi Lin, Ramakrishna Vedantam, Saurabh Gupta, Piotr
  Doll{\'a}r, and C~Lawrence Zitnick.
\newblock Microsoft coco captions: Data collection and evaluation server.
\newblock \emph{arXiv preprint arXiv:1504.00325}, 2015.

\bibitem[Chen et~al.(2020)Chen, Li, Yu, El~Kholy, Ahmed, Gan, Cheng, and
  Liu]{chen2020uniter}
Yen-Chun Chen, Linjie Li, Licheng Yu, Ahmed El~Kholy, Faisal Ahmed, Zhe Gan, Yu
  Cheng, and Jingjing Liu.
\newblock Uniter: Universal image-text representation learning.
\newblock In \emph{European conference on computer vision}, pages 104--120.
  Springer, 2020.

\bibitem[Cho et~al.(2021)Cho, Lei, Tan, and Bansal]{cho2021unifying}
Jaemin Cho, Jie Lei, Hao Tan, and Mohit Bansal.
\newblock Unifying vision-and-language tasks via text generation.
\newblock In \emph{International Conference on Machine Learning}, pages
  1931--1942. PMLR, 2021.

\bibitem[Chung et~al.(2022)Chung, Hou, Longpre, Zoph, Tay, Fedus, Li, Wang,
  Dehghani, Brahma, et~al.]{chung2022scaling}
Hyung~Won Chung, Le Hou, Shayne Longpre, Barret Zoph, Yi Tay, William Fedus,
  Yunxuan Li, Xuezhi Wang, Mostafa Dehghani, Siddhartha Brahma, et~al.
\newblock Scaling instruction-finetuned language models.
\newblock \emph{arXiv preprint arXiv:2210.11416}, 2022.

\bibitem[Cubuk et~al.(2020)Cubuk, Zoph, Shlens, and Le]{cubuk2020randaugment}
Ekin~D Cubuk, Barret Zoph, Jonathon Shlens, and Quoc~V Le.
\newblock Randaugment: Practical automated data augmentation with a reduced
  search space.
\newblock In \emph{Proceedings of the IEEE/CVF conference on computer vision
  and pattern recognition workshops}, pages 702--703, 2020.

\bibitem[Dai et~al.(2023)Dai, Li, Li, Tiong, Zhao, Wang, Li, Fung, and
  Hoi]{dai2023instructblip}
Wenliang Dai, Junnan Li, Dongxu Li, Anthony Meng~Huat Tiong, Junqi Zhao,
  Weisheng Wang, Boyang Li, Pascale Fung, and Steven Hoi.
\newblock Instructblip: Towards general-purpose vision-language models with
  instruction tuning, 2023.

\bibitem[Doveh et~al.(2023)Doveh, Arbelle, Harary, Schwartz, Herzig, Giryes,
  Feris, Panda, Ullman, and Karlinsky]{doveh2023teaching}
Sivan Doveh, Assaf Arbelle, Sivan Harary, Eli Schwartz, Roei Herzig, Raja
  Giryes, Rogerio Feris, Rameswar Panda, Shimon Ullman, and Leonid Karlinsky.
\newblock Teaching structured vision \& language concepts to vision \& language
  models.
\newblock In \emph{Proceedings of the IEEE/CVF Conference on Computer Vision
  and Pattern Recognition}, pages 2657--2668, 2023.

\bibitem[Fang et~al.(2023)Fang, Wang, Xie, Sun, Wu, Wang, Huang, Wang, and
  Cao]{fang2023eva}
Yuxin Fang, Wen Wang, Binhui Xie, Quan Sun, Ledell Wu, Xinggang Wang, Tiejun
  Huang, Xinlong Wang, and Yue Cao.
\newblock Eva: Exploring the limits of masked visual representation learning at
  scale.
\newblock In \emph{Proceedings of the IEEE/CVF Conference on Computer Vision
  and Pattern Recognition}, pages 19358--19369, 2023.

\bibitem[Gao et~al.(2023)Gao, Han, Zhang, Lin, Geng, Zhou, Zhang, Lu, He, Yue,
  et~al.]{gao2023llama}
Peng Gao, Jiaming Han, Renrui Zhang, Ziyi Lin, Shijie Geng, Aojun Zhou, Wei
  Zhang, Pan Lu, Conghui He, Xiangyu Yue, et~al.
\newblock Llama-adapter v2: Parameter-efficient visual instruction model.
\newblock \emph{arXiv preprint arXiv:2304.15010}, 2023.

\bibitem[Goyal et~al.(2017)Goyal, Khot, Summers-Stay, Batra, and
  Parikh]{goyal2017making}
Yash Goyal, Tejas Khot, Douglas Summers-Stay, Dhruv Batra, and Devi Parikh.
\newblock Making the v in vqa matter: Elevating the role of image understanding
  in visual question answering.
\newblock In \emph{Proceedings of the IEEE conference on computer vision and
  pattern recognition}, pages 6904--6913, 2017.

\bibitem[Honnibal and Montani(2017)]{spacy2}
Matthew Honnibal and Ines Montani.
\newblock {spaCy 2}: Natural language understanding with {B}loom embeddings,
  convolutional neural networks and incremental parsing.
\newblock To appear, 2017.

\bibitem[Hudson and Manning(2019)]{hudson2019gqa}
Drew~A Hudson and Christopher~D Manning.
\newblock Gqa: A new dataset for real-world visual reasoning and compositional
  question answering.
\newblock In \emph{Proceedings of the IEEE/CVF conference on computer vision
  and pattern recognition}, pages 6700--6709, 2019.

\bibitem[Jia et~al.(2021)Jia, Yang, Xia, Chen, Parekh, Pham, Le, Sung, Li, and
  Duerig]{jia2021scaling}
Chao Jia, Yinfei Yang, Ye Xia, Yi-Ting Chen, Zarana Parekh, Hieu Pham, Quoc Le,
  Yun-Hsuan Sung, Zhen Li, and Tom Duerig.
\newblock Scaling up visual and vision-language representation learning with
  noisy text supervision.
\newblock In \emph{International conference on machine learning}, pages
  4904--4916. PMLR, 2021.

\bibitem[Krishna et~al.(2017)Krishna, Zhu, Groth, Johnson, Hata, Kravitz, Chen,
  Kalantidis, Li, Shamma, et~al.]{krishna2017visual}
Ranjay Krishna, Yuke Zhu, Oliver Groth, Justin Johnson, Kenji Hata, Joshua
  Kravitz, Stephanie Chen, Yannis Kalantidis, Li-Jia Li, David~A Shamma, et~al.
\newblock Visual genome: Connecting language and vision using crowdsourced
  dense image annotations.
\newblock \emph{International journal of computer vision}, 123\penalty0
  (1):\penalty0 32--73, 2017.

\bibitem[Li et~al.(2023{\natexlab{a}})Li, Li, Le, Wang, Savarese, and
  Hoi]{li-etal-2023-lavis}
Dongxu Li, Junnan Li, Hung Le, Guangsen Wang, Silvio Savarese, and Steven~C.H.
  Hoi.
\newblock {LAVIS}: A one-stop library for language-vision intelligence.
\newblock In \emph{Proceedings of the 61st Annual Meeting of the Association
  for Computational Linguistics (Volume 3: System Demonstrations)}, pages
  31--41, Toronto, Canada, 2023{\natexlab{a}}. Association for Computational
  Linguistics.

\bibitem[Li et~al.(2022)Li, Li, Xiong, and Hoi]{li2022blip}
Junnan Li, Dongxu Li, Caiming Xiong, and Steven Hoi.
\newblock Blip: Bootstrapping language-image pre-training for unified
  vision-language understanding and generation.
\newblock In \emph{International Conference on Machine Learning}, pages
  12888--12900. PMLR, 2022.

\bibitem[Li et~al.(2023{\natexlab{b}})Li, Li, Savarese, and Hoi]{li2023blip2}
Junnan Li, Dongxu Li, Silvio Savarese, and Steven Hoi.
\newblock Blip-2: Bootstrapping language-image pre-training with frozen image
  encoders and large language models.
\newblock \emph{arXiv preprint arXiv:2301.12597}, 2023{\natexlab{b}}.

\bibitem[Li et~al.(2020)Li, Yin, Li, Zhang, Hu, Zhang, Wang, Hu, Dong, Wei,
  et~al.]{li2020oscar}
Xiujun Li, Xi Yin, Chunyuan Li, Pengchuan Zhang, Xiaowei Hu, Lei Zhang, Lijuan
  Wang, Houdong Hu, Li Dong, Furu Wei, et~al.
\newblock Oscar: Object-semantics aligned pre-training for vision-language
  tasks.
\newblock In \emph{Computer Vision--ECCV 2020: 16th European Conference,
  Glasgow, UK, August 23--28, 2020, Proceedings, Part XXX 16}, pages 121--137.
  Springer, 2020.

\bibitem[Liu et~al.(2023{\natexlab{a}})Liu, Emerson, and Collier]{liu2023vsr}
Fangyu Liu, Guy Emerson, and Nigel Collier.
\newblock Visual spatial reasoning.
\newblock \emph{Transactions of the Association for Computational Linguistics},
  11:\penalty0 635--651, 2023{\natexlab{a}}.

\bibitem[Liu et~al.(2023{\natexlab{b}})Liu, Li, Wu, and Lee]{liu2023visual}
Haotian Liu, Chunyuan Li, Qingyang Wu, and Yong~Jae Lee.
\newblock Visual instruction tuning.
\newblock \emph{arXiv preprint arXiv:2304.08485}, 2023{\natexlab{b}}.

\bibitem[Loshchilov and Hutter(2018)]{loshchilov2018decoupled}
Ilya Loshchilov and Frank Hutter.
\newblock Decoupled weight decay regularization.
\newblock In \emph{International Conference on Learning Representations}, 2018.

\bibitem[Luo et~al.(2023)Luo, Zhou, Ren, Chen, Sun, and Ji]{luo2023cheap}
Gen Luo, Yiyi Zhou, Tianhe Ren, Shengxin Chen, Xiaoshuai Sun, and Rongrong Ji.
\newblock Cheap and quick: Efficient vision-language instruction tuning for
  large language models.
\newblock \emph{arXiv preprint arXiv:2305.15023}, 2023.

\bibitem[Marino et~al.(2019)Marino, Rastegari, Farhadi, and
  Mottaghi]{marino2019ok}
Kenneth Marino, Mohammad Rastegari, Ali Farhadi, and Roozbeh Mottaghi.
\newblock Ok-vqa: A visual question answering benchmark requiring external
  knowledge.
\newblock In \emph{Proceedings of the IEEE/cvf conference on computer vision
  and pattern recognition}, pages 3195--3204, 2019.

\bibitem[Nivre et~al.(2016)Nivre, De~Marneffe, Ginter, Goldberg, Hajic,
  Manning, McDonald, Petrov, Pyysalo, Silveira, et~al.]{nivre2016universal}
Joakim Nivre, Marie-Catherine De~Marneffe, Filip Ginter, Yoav Goldberg, Jan
  Hajic, Christopher~D Manning, Ryan McDonald, Slav Petrov, Sampo Pyysalo,
  Natalia Silveira, et~al.
\newblock Universal dependencies v1: A multilingual treebank collection.
\newblock In \emph{Proceedings of the Tenth International Conference on
  Language Resources and Evaluation (LREC'16)}, pages 1659--1666, 2016.

\bibitem[OpenAI(2023)]{openai2023gpt}
R OpenAI.
\newblock Gpt-4 technical report.
\newblock \emph{arXiv}, pages 2303--08774, 2023.

\bibitem[Ouyang et~al.(2022)Ouyang, Wu, Jiang, Almeida, Wainwright, Mishkin,
  Zhang, Agarwal, Slama, Ray, et~al.]{ouyang2022training}
Long Ouyang, Jeffrey Wu, Xu Jiang, Diogo Almeida, Carroll Wainwright, Pamela
  Mishkin, Chong Zhang, Sandhini Agarwal, Katarina Slama, Alex Ray, et~al.
\newblock Training language models to follow instructions with human feedback.
\newblock \emph{Advances in Neural Information Processing Systems},
  35:\penalty0 27730--27744, 2022.

\bibitem[Radford et~al.(2021)Radford, Kim, Hallacy, Ramesh, Goh, Agarwal,
  Sastry, Askell, Mishkin, Clark, et~al.]{radford2021learning}
Alec Radford, Jong~Wook Kim, Chris Hallacy, Aditya Ramesh, Gabriel Goh,
  Sandhini Agarwal, Girish Sastry, Amanda Askell, Pamela Mishkin, Jack Clark,
  et~al.
\newblock Learning transferable visual models from natural language
  supervision.
\newblock In \emph{International conference on machine learning}, pages
  8748--8763. PMLR, 2021.

\bibitem[Rohrbach et~al.(2018)Rohrbach, Hendricks, Burns, Darrell, and
  Saenko]{rohrbach2018object}
Anna Rohrbach, Lisa~Anne Hendricks, Kaylee Burns, Trevor Darrell, and Kate
  Saenko.
\newblock Object hallucination in image captioning.
\newblock \emph{arXiv preprint arXiv:1809.02156}, 2018.

\bibitem[Saikh et~al.(2022)Saikh, Ghosal, Mittal, Ekbal, and
  Bhattacharyya]{saikh2022scienceqa}
Tanik Saikh, Tirthankar Ghosal, Amish Mittal, Asif Ekbal, and Pushpak
  Bhattacharyya.
\newblock Scienceqa: A novel resource for question answering on scholarly
  articles.
\newblock \emph{International Journal on Digital Libraries}, 23\penalty0
  (3):\penalty0 289--301, 2022.

\bibitem[Tan and Bansal(2019)]{tan2019lxmert}
Hao Tan and Mohit Bansal.
\newblock Lxmert: Learning cross-modality encoder representations from
  transformers.
\newblock \emph{arXiv preprint arXiv:1908.07490}, 2019.

\bibitem[Wang et~al.(2022)Wang, Yang, Men, Lin, Bai, Li, Ma, Zhou, Zhou, and
  Yang]{wang2022ofa}
Peng Wang, An Yang, Rui Men, Junyang Lin, Shuai Bai, Zhikang Li, Jianxin Ma,
  Chang Zhou, Jingren Zhou, and Hongxia Yang.
\newblock Ofa: Unifying architectures, tasks, and modalities through a simple
  sequence-to-sequence learning framework.
\newblock In \emph{International Conference on Machine Learning}, pages
  23318--23340. PMLR, 2022.

\bibitem[Wei and Zou(2019)]{wei2019eda}
Jason Wei and Kai Zou.
\newblock Eda: Easy data augmentation techniques for boosting performance on
  text classification tasks.
\newblock \emph{arXiv preprint arXiv:1901.11196}, 2019.

\bibitem[Xie et~al.(2020)Xie, Dai, Hovy, Luong, and Le]{xie2020unsupervised}
Qizhe Xie, Zihang Dai, Eduard Hovy, Thang Luong, and Quoc Le.
\newblock Unsupervised data augmentation for consistency training.
\newblock \emph{Advances in neural information processing systems},
  33:\penalty0 6256--6268, 2020.

\bibitem[Yang et~al.(2022)Yang, Gan, Wang, Hu, Ahmed, Liu, Lu, and
  Wang]{yang2022unitab}
Zhengyuan Yang, Zhe Gan, Jianfeng Wang, Xiaowei Hu, Faisal Ahmed, Zicheng Liu,
  Yumao Lu, and Lijuan Wang.
\newblock Unitab: Unifying text and box outputs for grounded vision-language
  modeling.
\newblock In \emph{European Conference on Computer Vision}, pages 521--539.
  Springer, 2022.

\bibitem[Yu et~al.(2023)Yu, Yang, Li, Wang, Lin, Liu, Wang, and Wang]{yu2023mm}
Weihao Yu, Zhengyuan Yang, Linjie Li, Jianfeng Wang, Kevin Lin, Zicheng Liu,
  Xinchao Wang, and Lijuan Wang.
\newblock Mm-vet: Evaluating large multimodal models for integrated
  capabilities.
\newblock \emph{arXiv preprint arXiv:2308.02490}, 2023.

\bibitem[Yuksekgonul et~al.(2022)Yuksekgonul, Bianchi, Kalluri, Jurafsky, and
  Zou]{yuksekgonul2022and}
Mert Yuksekgonul, Federico Bianchi, Pratyusha Kalluri, Dan Jurafsky, and James
  Zou.
\newblock When and why vision-language models behave like bags-of-words, and
  what to do about it?
\newblock \emph{arXiv e-prints}, pages arXiv--2210, 2022.

\bibitem[Yun et~al.(2019)Yun, Han, Oh, Chun, Choe, and Yoo]{yun2019cutmix}
Sangdoo Yun, Dongyoon Han, Seong~Joon Oh, Sanghyuk Chun, Junsuk Choe, and
  Youngjoon Yoo.
\newblock Cutmix: Regularization strategy to train strong classifiers with
  localizable features.
\newblock In \emph{Proceedings of the IEEE/CVF international conference on
  computer vision}, pages 6023--6032, 2019.

\bibitem[Zhang et~al.(2017)Zhang, Cisse, Dauphin, and
  Lopez-Paz]{zhang2017mixup}
Hongyi Zhang, Moustapha Cisse, Yann~N Dauphin, and David Lopez-Paz.
\newblock mixup: Beyond empirical risk minimization.
\newblock \emph{arXiv preprint arXiv:1710.09412}, 2017.

\bibitem[Zhang et~al.(2021)Zhang, Li, Hu, Yang, Zhang, Wang, Choi, and
  Gao]{zhang2021vinvl}
Pengchuan Zhang, Xiujun Li, Xiaowei Hu, Jianwei Yang, Lei Zhang, Lijuan Wang,
  Yejin Choi, and Jianfeng Gao.
\newblock Vinvl: Making visual representations matter in vision-language
  models.
\newblock \emph{arXiv preprint arXiv:2101.00529}, 1\penalty0 (6):\penalty0 8,
  2021.

\bibitem[Zhang et~al.(2022)Zhang, Roller, Goyal, Artetxe, Chen, Chen, Dewan,
  Diab, Li, Lin, et~al.]{zhang2022opt}
Susan Zhang, Stephen Roller, Naman Goyal, Mikel Artetxe, Moya Chen, Shuohui
  Chen, Christopher Dewan, Mona Diab, Xian Li, Xi~Victoria Lin, et~al.
\newblock Opt: Open pre-trained transformer language models.
\newblock \emph{arXiv preprint arXiv:2205.01068}, 2022.

\bibitem[Zheng et~al.(2023)Zheng, Chiang, Sheng, Zhuang, Wu, Zhuang, Lin, Li,
  Li, Xing, Zhang, Gonzalez, and Stoica]{zheng2023judging}
Lianmin Zheng, Wei-Lin Chiang, Ying Sheng, Siyuan Zhuang, Zhanghao Wu, Yonghao
  Zhuang, Zi Lin, Zhuohan Li, Dacheng Li, Eric.~P Xing, Hao Zhang, Joseph~E.
  Gonzalez, and Ion Stoica.
\newblock Judging llm-as-a-judge with mt-bench and chatbot arena, 2023.

\bibitem[Zhou et~al.(2022)Zhou, Yang, Loy, and Liu]{zhou2022conditional}
Kaiyang Zhou, Jingkang Yang, Chen~Change Loy, and Ziwei Liu.
\newblock Conditional prompt learning for vision-language models.
\newblock In \emph{Proceedings of the IEEE/CVF Conference on Computer Vision
  and Pattern Recognition}, pages 16816--16825, 2022.

\bibitem[Zhu et~al.(2023)Zhu, Chen, Shen, Li, and Elhoseiny]{zhu2023minigpt}
Deyao Zhu, Jun Chen, Xiaoqian Shen, Xiang Li, and Mohamed Elhoseiny.
\newblock Minigpt-4: Enhancing vision-language understanding with advanced
  large language models.
\newblock \emph{arXiv preprint arXiv:2304.10592}, 2023.

\end{thebibliography}
}

\newpage
\newpage
\appendix
\appendix


We present a comprehensive breakdown of the task instruction and ground truth answer templates for the \T~task, the essential components in constructing the caption correction task for model training. Additionally, we offer an extended set of data samples, showcasing the \T~task through its construction process.

\section{Templates for Task Construction}
The comprehensive template for the task construction (Section 4.3) is illustrated in Table~\ref{tab:temp}. To mitigate potential overfitting to the language bias inherent in the correction task, we employ various prompts and answer formats during data construction, as informed by empirical experiments. 
Specifically, for the BLIP-2 VLM models~\cite{li2022blip}, the shortest instruction, ``check the caption," is utilized. 
In the case of instructBLIP~\cite{dai2023instructblip} coupled with the Vicuna Large Language Model (LLM), multiple instructions are employed during tuning to prevent the degradation of generalization. 
With the different choice of these templates, it shows considerable effectiveness in training empirically.

\begin{table}[t]
\begin{center}
\resizebox{0.49\textwidth}{!}{ 
\begin{tabular}{l|l}
\toprule
\textbf{Type}                     & \textbf{Content}     \\ \midrule
\multirow{3}{*}{Task Instruction} & Check the caption: ``\{\}"                              \\
                                  & Check the caption according to the image: ``\{\}"       \\
                                  & Based on the image, please correct the caption: ``\{\}" \\ \midrule
\multirow{4}{*}{\textsf{\small{replace}} Answer}   & ``\{\}" should be ``\{\}"                               \\
                                  & ``\{\}" could be ``\{\}"                                \\
                                  & ``\{\}" is ``\{\}"                                      \\
                                  & ``\{\}" actually is ``\{\}''                            \\ \midrule
\multirow{4}{*}{\textsf{\small{swap}} Answer}      & ``\{\}" and ``\{\}" are swapped                         \\
                                  & ``\{\}" and ``\{\}" need to switch                      \\
                                  & ``\{\}" and ``\{\}" should exchange positions           \\
                                  & ``\{\}" and ``\{\}" need to be swapped \\  \bottomrule          
\end{tabular}
}
\end{center}
\caption{\textbf{The instruction prompts and answer templates for the ICCC task.} The ``\{\}" in each task instruction is a placeholder for augmented caption; the ``\{\}" in each answer template is the mismatched concept and the correct one, respectively.
}
\label{tab:temp}
\end{table}

\section{Example of Generated Samples}

\begin{figure*}[h]
    \centering
    \includegraphics[width=0.87\textwidth]{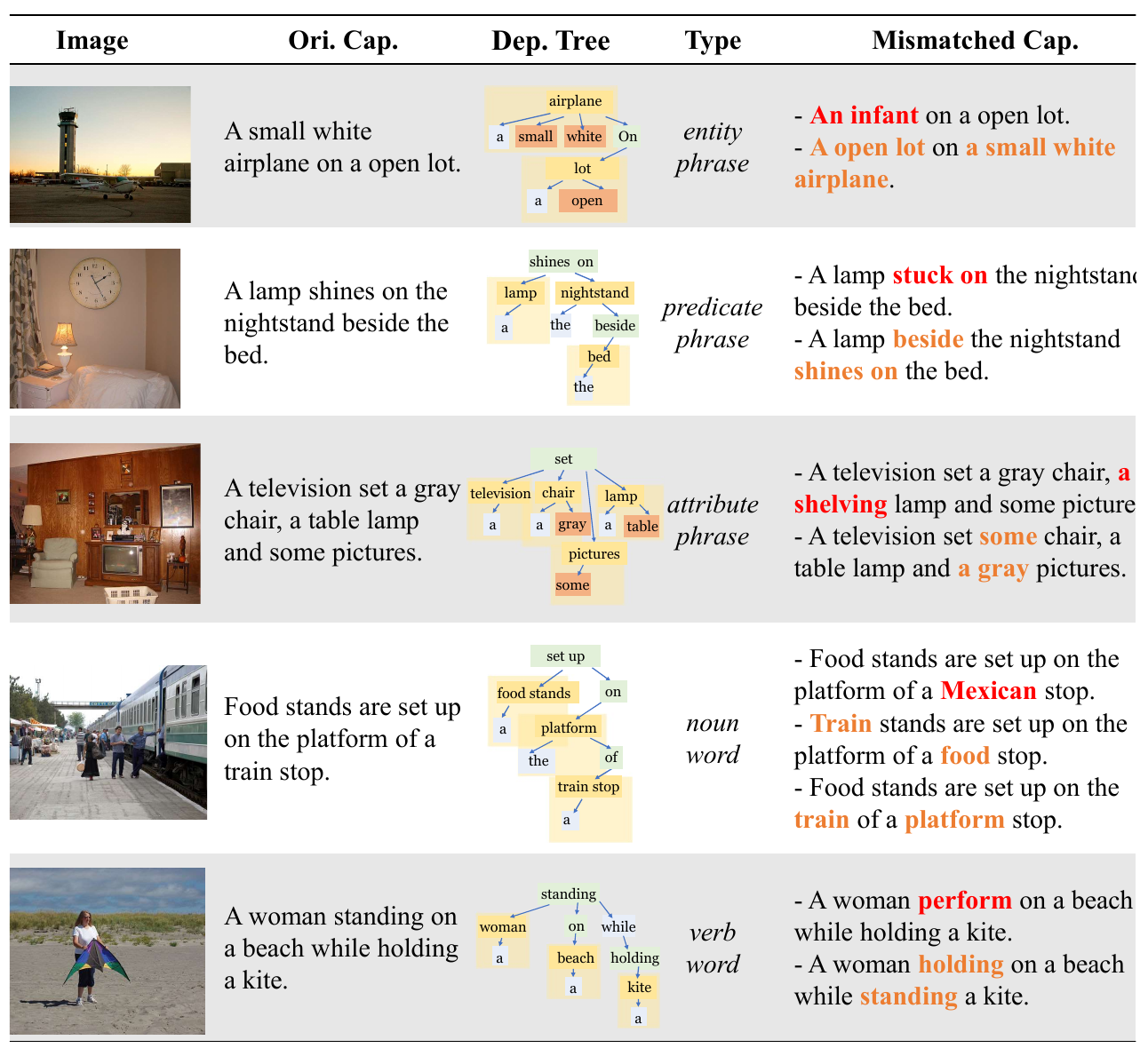}
    \caption{
        \textbf{Examples of constructed mismatched captions categorized by modification concept type.} 
        The linguistic units operated by \textsf{\small{replace}} are highlighted in \textcolor[RGB]{255,0,0}{\textbf{red}}, while the language units operated by \textsf{\small{swap}} are highlighted in \textcolor[RGB]{237,125,49}{\textbf{orange}}.
    }
    \label{fig:example_fig}
\end{figure*}

To vividly illustrate the data samples generated by our ICCC framework, we present examples of mismatched captions in Fig.~\ref{fig:example_fig}. Specifically, we showcase five examples, highlighting all concept types and the augmentation methods applied.

As illustrated in Fig.~\ref{fig:example_fig}, we parse the original caption (Ori. Cap) into a dependency tree (Dep. Tree), enabling the extraction of linguistic units associated with each concept type. 
Subsequently, specific types of linguistic units undergo replacement (\textsf{\small{replace}}), where affected elements are highlighted in \textcolor[RGB]{255,0,0}{\textbf{red}}, and swapping (\textsf{\small{swap}}) operations, with modified components emphasized in \textcolor[RGB]{237,125,49}{\textbf{orange}}. 
The operating results are the mismatched captions (Mismatched Cap.) with perturbed linguistic units, serving as input for VLMs to correct in the ICCC task.

Given that each caption involves multiple linguistic units for augmentation, the task constructor adeptly produces a variety of captions with mismatched concepts. Consequently, this ICCC task significantly enhances the image-conditioned text generation of VLMs, focusing on a general semantic concept, thereby effectively improving zero-shot capabilities across various downstream Vision-Language tasks.

\section{More Experiment Results} 

\subsection{Performance on Other VLMs} 
Our choice of VLMs is representative of the latest generative VLM architectures, which share similar structures by incorporating an adaptor to bridge the visual encoder and large language model. Notably, at the time of submission, InstructBLIP performed as the SoTA among the published generative VLMs, surpassing the zero-shot performance of Flamingo. Therefore, we selected InstructBLIP as a representative SoTA model for our experiments in the paper.
To demonstrate the efficacy of our approach with the latest and different models, we conducted experiments on the latest accepted SoTA method, LLaVA, published on NeurIPS 2023.
We follow the evaluation setup in the latest version, LLaVA-1.5, but exclude datasets used during instruction tuning, such as GQA and VQAv2, to maintain a zero-shot setting.
As shown in Tab.~\ref{tab:vlm}, our method outperforms the baseline model on two evaluation benchmarks by keeping consistency with the method setup outlined in the main text, even without adjusting hyperparameters.
This underscores the complementarity of our approach with existing instruction tuning methods.

\begin{table}[h]
\centering
\resizebox{0.76\columnwidth}{!}{
\begin{tabular}{l|cc}
\toprule
\textbf{LLaVA} & ScienceQA-IMG & MM-VET \\ \midrule
 Vicuna-7B      & 66.8                   &   30.8                  \\
 Vicuna-7B w/ ICCC    & \textbf{67.7}          & \textbf{31.7}                   \\ \bottomrule
\end{tabular}
}
\caption{The zero-shot evaluation results on LLaVA-7B with our tuning method.}\label{tab:vlm}
\end{table}

\subsection{Hallucinations and VSR} 
We evaluate the effect of the method on hallucinations by CHAIR score \cite{rohrbach2018object} on the COCO dataset. The results indicate that our approach proves beneficial in addressing caption hallucinations.
On the VSR dataset, our method also leads to further improvements on BLIP-2.
The hallucinations of VLMs are a very interesting research question, and we will conduct further exploration in future work.

\begin{table}[t]
\centering
\resizebox{0.78\columnwidth}{!}{
\begin{tabular}{l|cc|c}
\toprule
\multirow{2}{*}{\textbf{BLIP-2}} & \multicolumn{2}{c|}{\textbf{COCO}} & \multirow{2}{*}{\textbf{VSR}} \\
                                & CHAIRs           & CHAIRi          &                               \\ \midrule
OPT 2.7B                        & 3.2              & 2.3             & \textbf{48.28}                         \\
OPT 2.7B w/ ICCC                 & \textbf{3.0}              & \textbf{2.1}             & 47.79                         \\ \midrule
OPT 6.7B                        & 3.2              & 2.2             & 48.53                         \\
OPT 6.7B w/ ICCC                 & \textbf{3.0}              & \textbf{2.1}             & \textbf{51.55}                         \\ \midrule
FlanT5XL                        & 2.3              & 1.6             & 63.42                         \\
FlanT5XL w/ ICCC                 & \textbf{2.1}              & \textbf{1.5}             & \textbf{64.24}                         \\ \bottomrule
\end{tabular}
}
\caption{The zero-shot evaluation results on BLIP-2 on CHAIR metrics for image caption hallucinations and performance of VSR.}\label{tab:hal}
\end{table}

\renewcommand{\thetable}{S\arabic{table}}
\renewcommand{\thefigure}{S\arabic{figure}}

\end{document}